\documentclass[letterpaper]{article} 
\usepackage{aaai2026}  
\usepackage{times}  
\usepackage{helvet}  
\usepackage{courier}  
\usepackage[hyphens]{url}  
\usepackage{graphicx} 
\usepackage{natbib}  
\usepackage{caption} 
\usepackage{algorithm}
\usepackage{algorithmic}

\usepackage{newfloat}
\usepackage{listings}
\DeclareCaptionStyle{ruled}{labelfont=normalfont,labelsep=colon,strut=off} 
\floatstyle{ruled}
\newfloat{listing}{tb}{lst}{}
\floatname{listing}{Listing}
\usepackage{url}            
\usepackage{booktabs}       
\usepackage{amsfonts}       
\usepackage{nicefrac}       
\usepackage{microtype}      
\usepackage{xcolor}         
\usepackage{pifont}
\usepackage{multirow}

\title{UVLM: Benchmarking Video Language Model for Underwater World Understanding}
\author{
    Xizhe Xue\textsuperscript{\rm 1}\equalcontrib, Yang Zhou\textsuperscript{\rm 1}\equalcontrib,
    Dawei Yan\textsuperscript{\rm 1}, Lijie Tao\textsuperscript{\rm 1}, Junjie Li\textsuperscript{\rm 1}, \\
    Ying Li\textsuperscript{\rm 1},
    Haokui Zhang\textsuperscript{\rm 1}\thanks{Corresponding author}, Rong Xiao\textsuperscript{\rm 2}}
\affiliations{
    \textsuperscript{\rm 1}Northwestern Polytechnical University\\
    \textsuperscript{\rm 2}Intellifusion Inc.\\
  
}

\begin{document}

\maketitle




 \begin{abstract}

Recently, video-language models (VidLMs) have gained widespread attention and adoption. However, existing works primarily focus on terrestrial scenarios, overlooking the highly demanding application needs of underwater observation. To overcome this gap, we introduce UVLM, an under water observation benchmark which is build through a collaborative approach combining human expertise and AI models. To ensure data quality, we have conducted in-depth considerations from multiple perspectives. First, to address the unique challenges of underwater environments, we selected videos that represent typical underwater challenges including light variations, water turbidity, and diverse viewing angles to construct the dataset. Second, to ensure data diversity, the dataset covers a wide range of frame rates, resolutions, 419 classes of marine animals, and various static plants and terrains. Next, for task diversity, we adopted a structured design where observation targets are categorized into two major classes: biological and environmental. Each category includes content observation and change/action observation, totaling 20 subtask types. Finally, we designed several challenging evaluation metrics to enable quantitative comparison and analysis of different methods. Experiments on two representative VidLMs demonstrate that fine-tuning VidLMs on UVLM significantly improves underwater world understanding while also showing potential for slight improvements on existing in-air VidLM benchmarks, such as VideoMME and Perception text. The dataset and prompt are publicly available at: \url{https://github.com/Cecilia-xue/UVLM-Benchmark}.

\end{abstract}
\section{Introduction}
\begin{figure*}
  \centering  
  \includegraphics[width=0.9\textwidth]{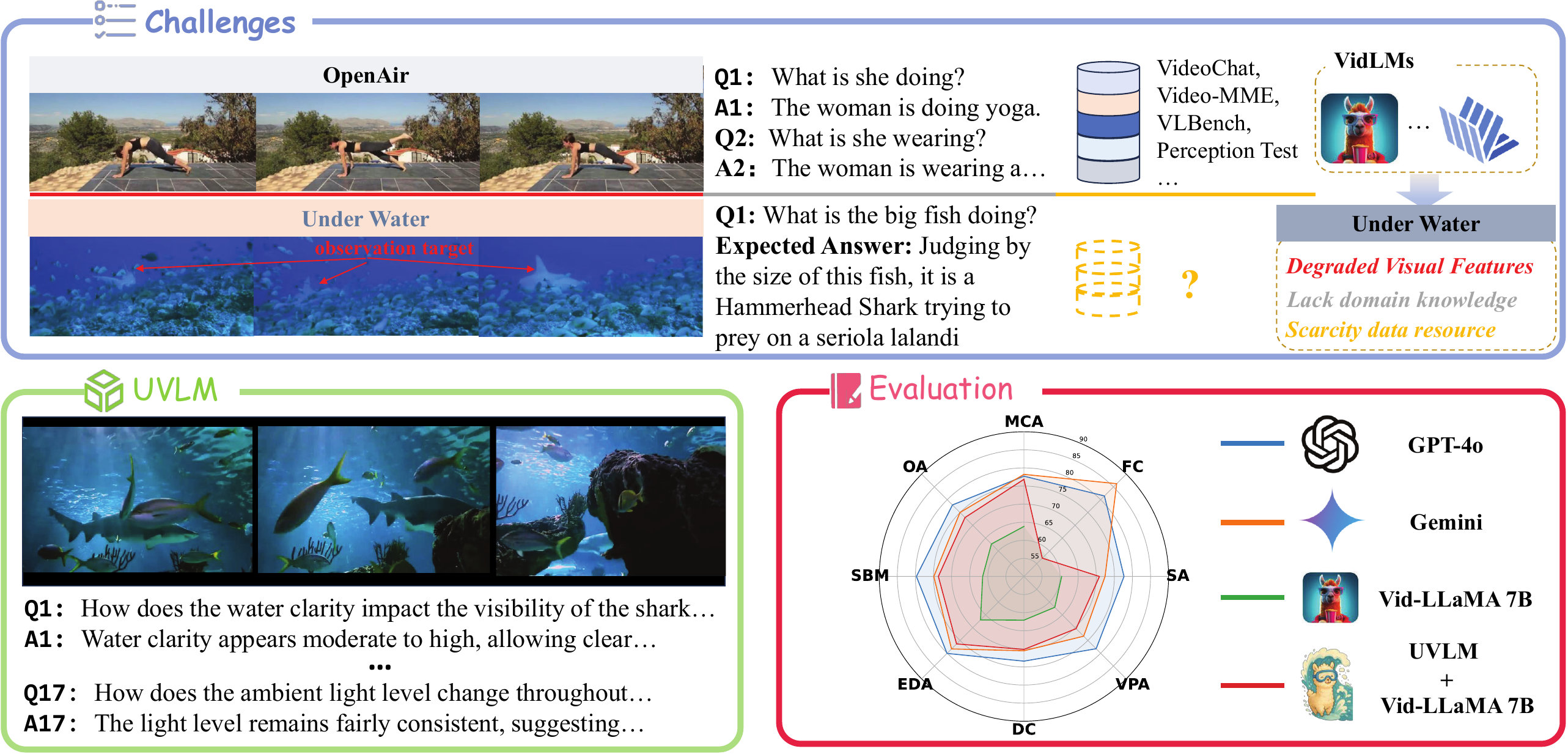}
  \caption{Challenges for VidLMs for understanding underwater videos. Our UVLM is proposed to overcome this gap and it enables the 7B VidLM to achieve performance comparable to closed-source models like GPT-4o and Gemini.}  
  \label{fig:abstract}
\end{figure*}

Video-language understanding~\cite{xu2016msr,anne2017localizing} stands at the forefront of multimedia research, empowering systems to interpret, reason about, and generate natural language descriptions of temporal and dynamic visual content~\cite{patraucean2023perception}. Recent advances~\cite{patraucean2023perception,li2024mvbench,zhang2025videollama} in video-language models (VidLMs) have demonstrated impressive performance in tasks such as video captioning, temporal grounding, and visual question answering, focusing primarily on human-centric scenarios and common object interactions. Despite these achievements, a critical question remains: \textbf{Can current VidLMs effectively understand videos captured in special imaging conditions, such as underwater environments?}

This question is especially relevant as underwater environments constitute an uncharted domain with immense scientific value (e.g., marine biodiversity monitoring, ecosystem health assessment~\cite{xue2024reo}) and substantial engineering applications (e.g., autonomous underwater vehicles, offshore infrastructure inspection). As illustrated in Figure \ref{fig:abstract}, applying VidLMs to underwater content presents three challenges that distinguish this domain from conventional video-language tasks:
\begin{itemize}
    \item \textbf{Degraded Visual Features Hindering Analytical Decisions.} The underwater domain introduces fundamental perceptual barriers that confound existing video understanding approaches. Underwater scenes exhibit variable illumination with rapid light attenuation at depth, wavelength-dependent color distortion that shifts the visual spectrum, fluctuating turbidity affecting visibility~\cite{islam2020fast, akkaynak2019sea}. Standard VidLMs designed for terrestrial scenario environments struggle to perform effectively underwater due to these low-quality visual cues.
    
    \item \textbf{Lack of Scientific Domain Knowledge.} Underwater content requires specialized ecological expertise for accurate interpretation. Unlike common scenarios featuring familiar objects and actions, underwater videos capture complex species interactions, specialized behavioral patterns, and environmental relationships that demand expert knowledge to decode~\cite{marks2022deep,han2024multi}. The interpretation of underwater footage requires understanding multiple layers of information, including taxonomic identification, morphological characteristics, behavioral states, and environmental contexts. Models must capture the intricate ecological relationships between marine organisms' appearance, behaviors, and habitats, creating a significant knowledge gap for systems trained primarily on common objects and human activities.
    
    \item \textbf{Data Resource Scarcity.} The development of effective video-language models for underwater environments is critically hindered by the absence of comprehensive training resources. Current video-language datasets predominantly focus on everyday human activities (HowTo100M~\cite{miech2019howto100m}), sports (Sports-1M~\cite{karpathy2014large}), or general-purpose actions (Kinetics~\cite{carreira2017quo}), offering minimal support for specialized scientific domains. While underwater-specific datasets exist, they typically prioritize narrow tasks such as object tracking (WebUOT~\cite{zhangwebuot}) or instance segmentation (UIIS~\cite{watermask}). These image-based datasets fail to equip models with scientific domain knowledge necessary for comprehensive understanding of marine organism behaviors in underwater videos. This data gap creates a major barrier to advancing video-language understanding beyond common scenarios to specialized scientific contexts.
\end{itemize}

To bridge these gaps, a comprehensive Underwater Video-Language Multimodal (UVLM) dataset with professional annotations is presented. Compared to existing video language understanding benchmarks, UVLM demonstrates distinct differences in both content composition and construction methodology. First, in terms of content, UVLM is the first video-language benchmark specifically designed for underwater environments. To ensure the dataset accurately captures the distinctive characteristics of underwater settings, we carefully selected videos that encompass unique challenges of this domain, including low-light conditions, water turbidity, and the highly variable movement patterns of marine organisms. Second, regarding construction methodology, we adopted a structured framework combining human-AI collaboration. The annotation targets encompass both biological and environmental elements, covering static and dynamic scenarios. After frame-by-frame manual annotation, we leveraged these human-annotated labels to guide GPT-4o in generating diverse sample content. Finally, all data underwent manual verification, with inconsistent entries either re-annotated or supplemented using search engines for factual accuracy. 

The final dataset comprises 0.9M frames, 419 distinct biological categories, and diverse underwater scenarios, encompassing 20 types of video-language understanding tasks. Additionally, we established eight specialized metrics for quantitative performance comparison on the dataset. UVLM advances video-language research through its comprehensive coverage of core technical challenges:
\begin{itemize}
    \item Temporal Understanding: Supports development of models that can interpret continuous behavior sequences and environmental changes over time.
     \item Fine-Grained Recognition: Enables research on distinguishing subtle visual differences with significant scientific meaning.
     \item Compositional Reasoning: Facilitates the development of models that can decompose complex scenes into scientifically meaningful components.
     \item  Knowledge-Grounded Generation: Provides a foundation for generating technically accurate language descriptions based on visual evidence. 
\end{itemize}

\begin{table*}
\centering
\begin{tabular}{lllllllllll}
\hline

Dataset             & Venue           & Im       & Vid     & Lang &OQA & BioInf& Seq & Frame  & Task          & Categ \\ 
\hline
LSUI \cite{LSUI}    & TIP$\prime$23   & \checkmark  & \ding{55}  &  \ding{55} & \ding{55}  & \ding{55}  & -        & 5k      & IR           & 10                \\
DRUVA \cite{DRUVA}   & ICCV$\prime$23             & \checkmark     & \checkmark     &  \ding{55} & \ding{55}  & \ding{55}        & 20       & 6K   & DE, IR & 20                \\
IOCfish5K \cite{sun2023indiscernible}    & CVPR$\prime$23             & \checkmark     &  \ding{55}       &  \ding{55}    &  \ding{55} & \ding{55}    & -       & 5K & OC             & -                 \\
UIIS \cite{watermask}     & ICCV$\prime$24             & \checkmark     & \ding{55}     & \ding{55}  &\ding{55} & \ding{55}       & -        & 4.6k     & IS          & 7                \\
VMAT \cite{VMAT}  & IJCV$\prime$23             & \checkmark     & \checkmark     & \ding{55}   &\ding{55} & \ding{55}      & 33      & 57K     & SOT                    & 17          \\
WebUOT \cite{zhangwebuot}  & NeurIPS$\prime$24             & \checkmark     & \checkmark     & \checkmark      &\ding{55} & \ding{55}    & 1500      & 1M     &SOT                    & 408    \\  
MarineInst \cite{zheng2024marineinst}  & ECCV$\prime$24             & \checkmark     &\ding{55}     & \checkmark      & \ding{55}  & \ding{55} &-        & 2.42 M     &IS, CAP                    & -    \\   
USOD \cite{USOD10K}  & TIP$\prime$25             & \checkmark     &\ding{55}   &\ding{55} & \ding{55}    & \ding{55}    &-        & 10K     &SOD                    & 70    \\    \hline
UVLM    & -                   & \checkmark     & \checkmark     & \checkmark       & \checkmark     & \checkmark       & 2111      & 0.86M    &    VU                                     & 419                \\ \hline

\end{tabular}
\caption{Comparison of recent underwater observation datasets. Im, Vid, Lang, OQA, BioInf, Seq, and Categ denote Image, Video, Language, Open-ended Question Answering, Taxonomic Classification Information in Biology, Sequence, and Category, respectively. In Task, IR, DE, OC, IS, SOT, CAP, SOD, and VU refer to Image Restoration, Depth Estimation, Object Counting, Instance Segmentation, Single Object Tracking, Captioning, Salient Object Detection, and Video Understanding, respectively.}
\label{tab:uwBenchmarks}
\end{table*}

In summary, UVLM represents a significant step toward extending the capabilities of video-language models beyond everyday scenarios to specialized scientific domains. By providing richly annotated video-language pairs in underwater environments, UVLM enables the development of models that can interpret complex ecological dynamics and communicate this understanding through natural language, ultimately contributing to both multimedia research and marine science.

\section{Related Work}
\label{sec:Related}


\subsection{Video Language Understanding Benchmark}

In recent years, many video language benchmarks have emerged, each targeting specific themes and application domains. Some focus primarily on everyday life scenarios ~\cite{xu2017video,xiao2021next,yu2019activitynet}, while others emphasize human action or movie clips ~\cite{mangalam2023egoschema,song2024moviechat}. More comprehensive datasets cover a wider range of categories, including knowledge, sports, and instructional videos ~\cite{li2024mvbench,wu2024longvideobench,fu2024video,wang2024lvbench}. Table 1 in Appendix.B provides a concise summary of representative benchmarks, highlighting their respective content coverage. Although these efforts have significantly advanced video-language research in terrestrial scenario contexts, the underwater environment remains largely uncharted, motivating our investigation of underwater video-language benchmark.

\subsection{Underwater Observation Datasets}

Underwater observation datasets and benchmarks have followed a clear progression, evolving from low-level tasks to high-level ones, from single-frame image analysis to video content analysis, and from unimodal to multimodal approaches. We summarize several key underwater image datasets and compare them with our UVLM in Table \ref{tab:uwBenchmarks}.

Low-level perception datasets focus on fundamental image enhancement and quality assessment, such as LSUI~\cite{LSUI}. These datasets provide essential resources for developing and validating algorithms to address the typical degradation in underwater imagery. More recent efforts have extended this work into the temporal domain with video datasets like DRUVA~\cite{DRUVA}, which contains 6,000 frames. Mid-level recognition tasks have also benefited from dedicated datasets. For instance, Wildfish~\cite{wildfish} is tailored for marine life recognition, while COU~\cite{COU} supports segmentation research. Additionally, a range of tracking datasets, such as VMAT~\cite{VMAT} and  WebUOT~\cite{zhangwebuot} has further advanced object-centric underwater understanding. At the high-level, multimodal tasks are beginning to emerge. The MarineInst dataset~\cite{zheng2024marineinst} marks a critical step forward by supporting advanced tasks such as image segmentation and captioning, thereby opening new avenues for comprehensive underwater analysis.

Although these advancements have significantly propelled multimodal underwater analysis, multimodal underwater video analysis remains largely unexplored. Underwater videos inherently capture rich temporal dynamics, such as marine life trajectories and multi-view morphological changes, which provide additional contextual cues. By jointly leveraging visual appearance, temporal evolution, and domain-specific textual knowledge, unique advantages emerge. This holistic approach holds considerable promise for enhancing marine life interpretation, ecological monitoring, and overall environmental understanding.


\begin{figure*}
    \centering  
    \includegraphics[width=0.9\textwidth]{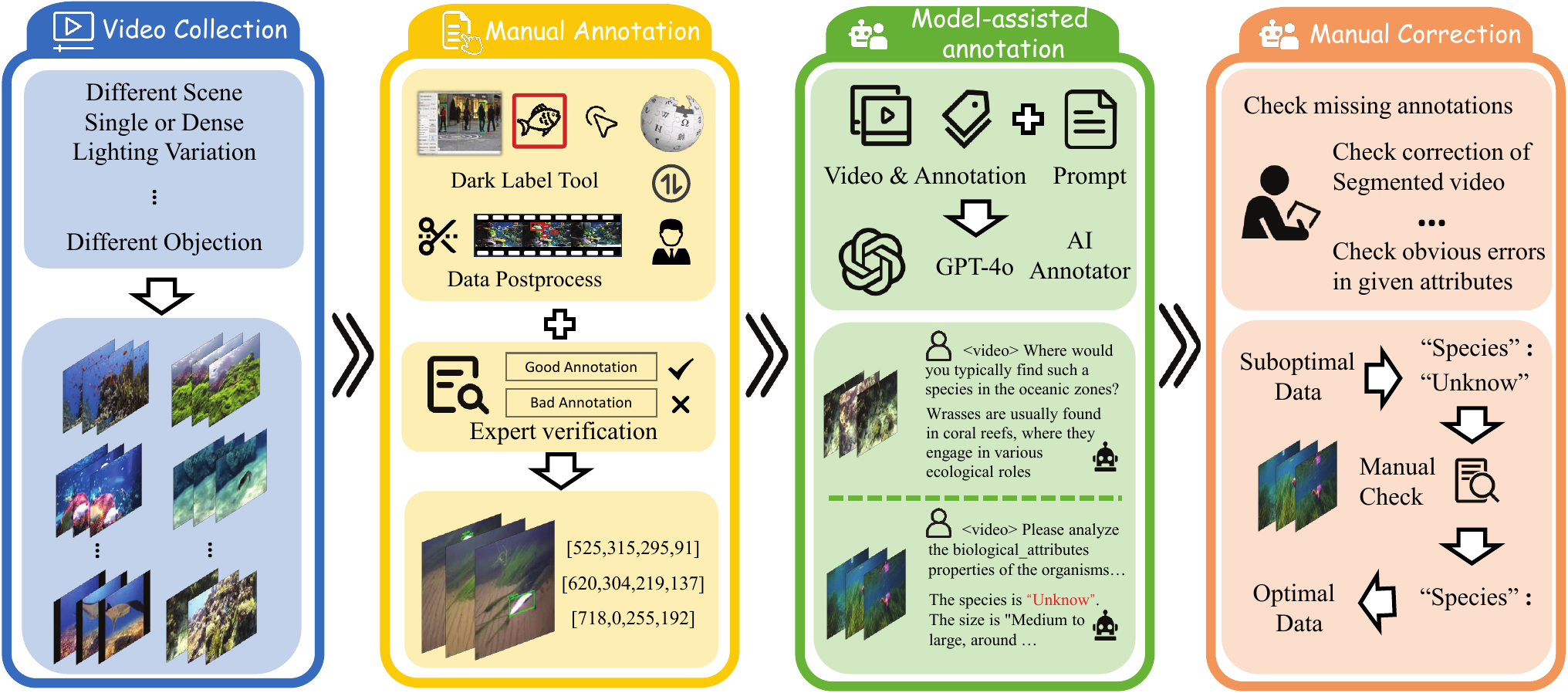} 
    \caption{An overview of data preparation and generation pipeline for UVLM. }
    \label{fig:data-pipeline}
\end{figure*}

\section{UVLM}
\label{sec:uvlm}


\subsection{Overview}
Currently, research on VidLMs and the construction of corresponding benchmarks has been ongoing for several years, achieving significant progress. However, existing work primarily focuses on land scenarios, neglecting underwater environments. Extending VidLM technology to the underwater domain still faces challenges such as visual feature degradation caused by harsh aquatic conditions, the lack of domain knowledge due to domain gaps, and the scarcity of benchmarks due to difficulties in data acquisition. 

In this paper, taking these challenges into consideration, we developed UVLM, the first underwater video-language benchmark. Firstly, we collect videos from typical underwater environments like oceans, lakes, and rivers, which exhibit different degradation problems such as water surface ripples, turbid water, and light scattering. Secondly, we select videos containing different content types, including static environmental features, static observations of organisms, dynamic scene variations, and dynamic biological behaviors, etc. 
To systematically inject domain knowledge into the dataset, we design textual descriptions structured into 20 subtasks, reflecting VidLMs' capabilities across 9 dimensions, such as marine animal behavior, water bodies, and geological features (details in Appendix.B). 
Thirdly, to maximize the scale of our dataset, we adopt a dual approach: systematically collecting data from online sources while strategically screening and re-annotating  existing relevant datasets. 

The final dataset comprises approximately two thousand carefully selected video sequences and 0.9 million frames, covering 419 different marine organisms and about 40 thousand video-language pairs. Previous underwater observation datasets~\cite{lian2023watermask,9416821,Alawode_2022_ACCV,alawode2023improvingunderwatervisualtracking,zhang2025underwatercamouflagedobjecttracking,fan2019lasothighqualitybenchmarklargescale,Huang_2021} typically focus on traditional computer vision tasks such as image segmentation and single object tracking, which are pure vision tasks. A few underwater observation datasets~\cite{zheng2024marineinst,LI2025440} also incorporate visual and language elements, but the tasks are still limited to typical applications such as captioning. Unlike these datasets, the proposed UVLM is designed to incorporate specific underwater domain knowledge and requirements. To achieve this, we designed 16 to 20 relative questions referring to underwater research topics such as marine organism recognition, behavioral analysis and prediction, habitat pattern characterization, subaquatic environmental monitoring, and ecosystem dynamics assessment, etc., covering both biological and environmental dimensions critical to underwater exploration and conservation. A comparison with previous underwater datasets is shown in Table \ref{tab:uwBenchmarks}.

\begin{figure*}
    \centering  
    \includegraphics[width=0.9\textwidth]{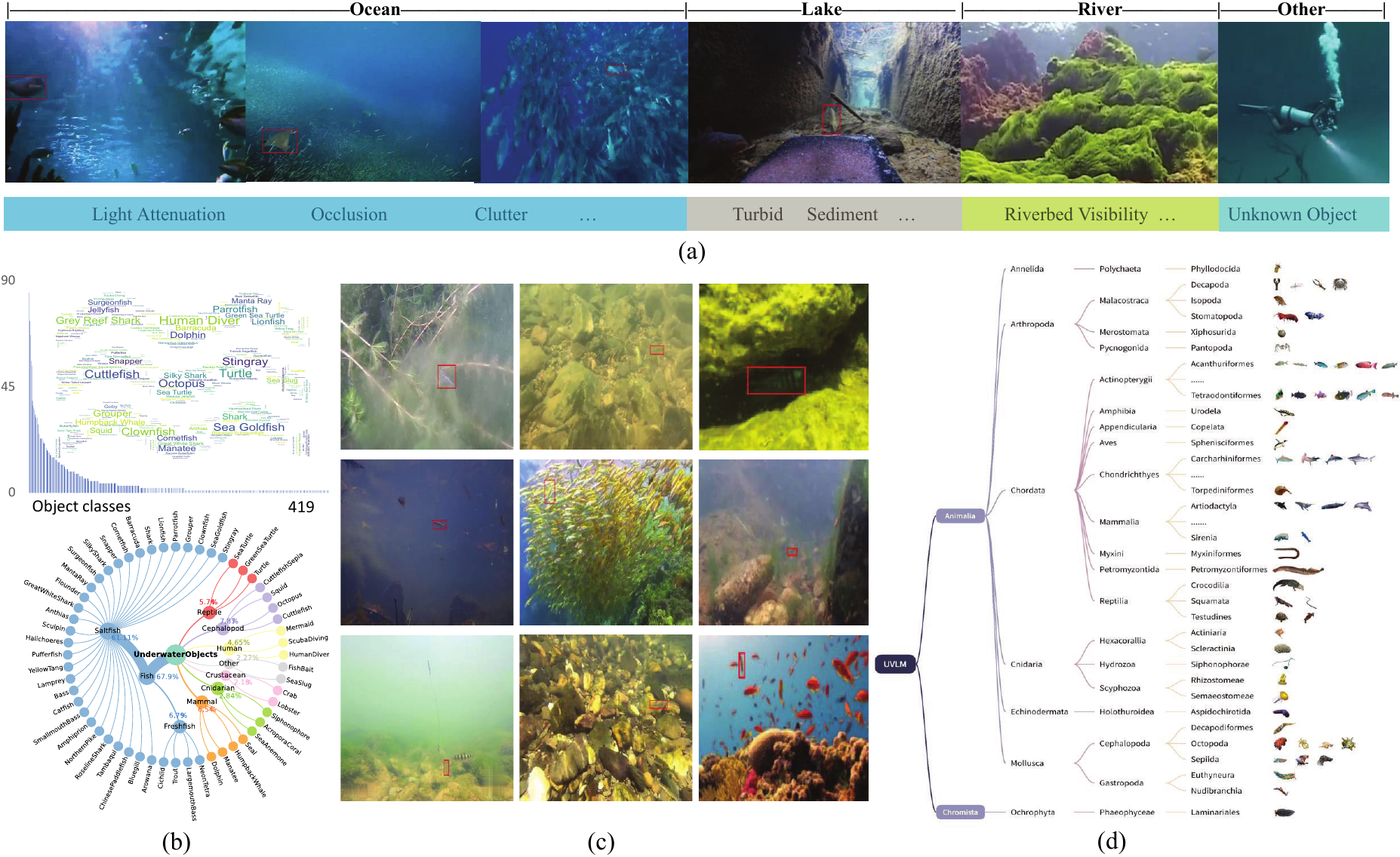} 
    \caption{Statistics on UVLM. (a) Scene distribution; (b)  
    Observation target distribution; (c) Several samples in UVLM; (d)  Fine-grained taxonomic classification information (partial categories), from left to right: Kingdom, Phylum, Class, Order.}.
    \label{fig:statistic}   
\end{figure*}

\subsection{Data Collection and Annotation}

As shown in Figure \ref{fig:data-pipeline}, the benchmark construction consists for four major parts: video collection, manual annotation, model-assisted annotation and manual correction. 

\noindent \textbf{Video collection}. Our objective extends beyond merely constructing an underwater video-language benchmark and training VidLMs on the data. Instead, we aim to build a dataset that captures the unique challenges of the underwater environment and enables VidLMs to operate in underwater contexts to support related research. To achieve this, video collection must satisfy two criteria: 1) the videos must capture distinctive underwater challenges; 2) the dataset must incorporate an adequate volume of representative cases to enable effective VidLM fine-tuning. To meet these requirements, we implement a dual-path video acquisition strategy. 

The first path is collecting underwater videos from websites such as youtube and bilibili. We collect web-crawled data while enforcing two quality control principles during acquisition: 1) \textit{Both camera and observation objects or scenes must be under water.} This selection rule is more aligned with real world underwater vehicle and robot application scenarios; 2) \textit{The typical underwater scenarios and targets with different challenges and characteristics must be covered.} 
The videos should cover typical environments, like oceans, lakes, rivers and fish tanks, etc. Meanwhile, the chosen targets should include (but not limited to) common animals (fish, whales, prawns, tortoises, etc.), people, mermaids and generic objects. With the principles, we selected about 400 video sequences, each of which contains 100 to 3000 frames, covering 53 classes of objects. The second path is re-annotating existing dataset. For scalability considerations, we selected WebUOT~\cite{zhangwebuot} as our primary data source.

\noindent \textbf{Manual annotation}. As emphasized in the Introduction, the substantial domain gap between terrestrial and underwater environments requires auxiliary annotation information to effectively leverage existing VidLMs for collaborative labeling. To facilitate granular analysis of underwater targets, such as studying the behavioral traits and ecological patterns of marine organisms, we annotate each frame with both bounding boxes and fine-grained taxonomic classifications aligned with marine biology standards. These annotations function as priors when fed into AI models for text generation.


For videos collected from websites, we employed 12 annotators annotate videos following three principles: 1) For occluded objects, only the visible part is marked with a rectangular box; 2) For the sharply protruding part, such as the tail and mouth of fish, whether to contain it in the bounding box is determined based on the proportion of the target and background in the additional introduced area. In specific, if the object’s protruding parts accounts for more than one-third of the additional introduced area, the sharply protruding part would be marked, otherwise it should not be marked; 3) Each frame is annotated with an axis aligned bounding box using the DarkLabel toolbox\footnote{https://github.com/darkpgmr/DarkLabel}. Each sequence is assessed by three domain experts to minimize annotation errors. Unqualified labels would be sent to other annotators for re-labeling. 

For videos from WebUOT, we followed several steps to refine video quantity. We first performed data cleaning to filter out videos with information interference, such as those containing large areas of subtitles or watermarks that affect video quality. Then, we conducted scene cleaning to remove videos that were not in natural underwater environments, such as those filmed in aquariums, fish tanks, or simulated gaming scenes. Next, we adopted SAM~\cite{kirillov2023segment} and LaMa~\cite{suvorov2022resolution} to remove small-area watermarks, subtitles, and other information interference. We use mouse clicks to select the area, SAM for segmentation, and then LaMa for inpainting. Next, to avoid challenges caused by excessively long videos and focus specifically on evaluating the impacts from underwater environments, We segmented some longer videos with fewer observed organisms into 300-600 frame clips to balance the sample distribution. Some videos collected from the internet even contain up to 3,000 frames, so the dataset can also evaluate the models' ability to understand extended underwater scenarios.

To achieve reliable and fine-grained taxonomic classification of marine animals in these videos, we implemented a structured three-phase annotation procedure. In the first phase, four annotators with extensive marine biology knowledge independently labeled each target within the video frames, providing both species-level identifications and detailed taxonomic classifications according to the classic five-kingdom system (kingdom, phylum, class, order, etc.) in biology~\cite{whittaker1969new}, supported by authoritative sources such as Wikipedia. In the second phase, annotations were cross-validated by annotator pairs. Any disagreements were resolved by consulting a third annotator for majority consensus, with unresolved cases flagged for expert review. In the third phase, a senior marine biology expert reviewed the validated annotations, marking questionable cases. Finally, all flagged annotations were collectively discussed by five experts to establish definitive classifications. This rigorous approach ensured high accuracy and consistency in fine-grained taxonomic annotation.

\begin{table*}
\centering
\begin{tabular} {c|cc|cccccc}
\hline
\multirow{2}{*}{\textbf{Method}} & 
\multicolumn{2}{c|}{\textbf{Objective Metrics}} & 
\multicolumn{6}{c}{\textbf{LLM-based Judgement Metrics}} \\
\cline{2-9}
 &

MCA     &FGC    &SA &DC &VPA    &EDA    &SBM    &Overall Accuracy \\ 
\hline
\multicolumn{9}{c}{Closed-source VidLMs and other Open Source Large VidLMs}  \\ 
\hline
GPT-4o                  & 77.72 & 81.47 & 77.67 & 73.40 & 78.23 & 80.07 & 79.73 & 77.95   \\ 
Claude3.7-Sonnet        & 76.61 & 82.64 & 73.35 & 73.58 & 74.10 & 79.71 & 76.35 & 76.09   \\
Gemini2.5-Flash         & 78.22 & 86.27 & 72.43 & 73.34 & 70.53 & 78.32 & 74.92 & 75.00   \\
Qwen2.5VL-72B           & 75.97 & 80.57 & 74.22 & 71.94 & 74.85 & 78.45 & 77.40 & 75.49   \\ 
\hline
\multicolumn{9}{c}{Base VLM} \\
\hline
Qvis2.5-2B              & 61.87 & 29.85 & 52.88 & 49.70 & 54.66 & 55.81 & 50.62 & 53.06   \\
Qvis2.5-2B + UVLM
                        & 72.89 & 59.25 & 67.64 & 62.39 & 65.30 & 69.05 & 66.29 & 68.85   (\textbf{+15.79}) \\
\hline
\multicolumn{9}{c}{Base VidLMs} \\
\hline
InternVL2.5-1B          & 48.54 & 29.61 & 44.35 & 47.30 & 45.30 & 50.15 & 44.85 & 46.73   \\
VideoLLaMA3-2B          & 57.17 & 31.25 & 55.89 & 58.48 & 58.53 & 63.15 & 55.74 & 58.39   \\

Qwen2.5VL-2B            & 59.47 & 35.23 & 54.12 & 50.65 & 53.47 & 57.31 & 50.60 & 52.97   \\
InternVL2.5-8B          & 57.64 & 36.63 & 57.70 & 59.75 & 59.45 & 63.48 & 61.05 & 60.15   \\
VideoLLaMA3-7B          & 63.83 & 42.62 & 60.43 & 62.07 & 62.05 & 66.97 & 61.41 & 62.70   \\
Qwen2.5VL-7B            & 66.22 & 48.36 & 66.67 & 59.98 & 61.74 & 68.41 & 61.79 & 63.57   \\
\hline
InternVL2.5-1B + UVLM
                       & 64.52 & 45.37 & 54.48 & 56.55 & 55.78 & 65.16 & 58.74 & 59.14 (\textbf{+12.41})   \\

VideoLLaMA3-2B + UVLM
                       & 70.41 & 46.35 & 63.54 & 64.76 & 66.71 & 70.28 & 67.32 & 66.67
                       (\textbf{+8.28}) \\
                       

Qwen2.5VL-2B + UVLM
                        & 62.38 & 56.08 & 60.47 & 55.81 & 57.93 & 61.45 & 58.26 & 58.44   (\textbf{+5.47}) \\
                        
InternVL2.5-8B + UVLM
                       & 70.26 & 43.94 & 65.66 & 66.42 & 65.05 & 71.10 & 69.98 & 69.45   (\textbf{+9.30}) \\

VideoLLaMA3-7B + UVLM
                       & 76.85 & 57.17 & 70.88 & 70.17 & 70.40 & 76.35 & 73.66 & 73.04   (\textbf{+10.34}) \\
                       
Qwen2.5VL-7B + UVLM
                        & 71.69 & 63.41 & 70.47 & 67.25 & 67.29 & 72.16 & 65.76 & 68.08  (\textbf{+4.51}) \\

\hline
\end{tabular}
\caption{Performance comparison of different methods on UVLM test set. Metric abbreviations: MCA, FGC, SA, DC, VPA, EDA, SBM denote Multiple Choice Accuracy, Fine-grained Taxonomic Classification, Semantic Accuracy, Detail Completeness, Visual Perception Accuracy, Environmental Description Accuracy, and Species Behavior Matching, respectively.}
\label{tab:comparison}
\end{table*}

\noindent \textbf{Model-assisted annotation}. After preparing the videos, we carefully designed prompts to guide GPT-4o in generating relevant questions and answers based on the input video content. To ensure the generated questions have practical domain relevance, we first conducted research on key topics in marine biology, such as observed species, organism behavior, and habitat characteristics covered in the videos. Guided by these findings, we designed prompts to instruct GPT-4o to generate content along two dimensions: 1) \textit{Marine Organism Dimension}. Static aspects: Species identification, biological attributes (e.g., morphology, coloration). Dynamic aspects: Behavioral analysis (e.g., feeding, interactions), movement patterns. 2) \textit{Underwater Environment Dimension}. Static aspects: Environmental features (e.g., substrate type, coral structures), habitat traits. Dynamic aspects: Light condition variations, visibility fluctuations, etc. 

The questions include two formats: multiple-choice and open-ended. For instance, multiple-choice: \textit{"What marine species is primarily observed in the images? A) Clownfish B) Chromis dimidiatus C) Parrotfish".} Open-ended: \textit{"What is the overall setting of the video, and how does it influence Chromis dimidiatus' activities?".} During the Q\&A process, we implemented clear prompt constraints, including: 1) Topic-specific constraints (e.g., focusing on taxonomic details, observed behaviors, or physicochemical properties of the water body); 2) Style requirements (e.g., diversifying sentence structures and enriching question types). Each video generates 16 to 20 video-text pairs. Appendix.A offers the generation prompts. 

\noindent \textbf{Manual correction}. Despite the integration of domain experts and model-assisted generation in our annotation pipeline, the intrinsic complexity of underwater scenes and the limitations of automatic generation inevitably introduce semantic drift and consistency issues. To guarantee the ecological validity and scientific rigour of the final dataset, we therefore incorporated a dedicated manual-correction stage as the decisive quality-assurance step.

Stringent quality control was applied to all Q\&A pairs generated by GPT-4o through a two-tier human review protocol. The first round, conducted by general reviewers, focused on detecting conflicts between the information supplied to GPT-4o and the content it produced. The second round involved senior experts performing in-depth edits to ensure factual precision and domain conformity. During both rounds, hallucinations or statements irrelevant to the video context were either removed or rewritten. Whenever a model description diverged from ground truth, we leveraged the accompanying assistant information to cross-validate and revise the erroneous content, safeguarding dataset integrity.

\subsection{UVLM Statistics}

The scene distribution of UVLM and the distribution of various observation targets are presented in Figure \ref{fig:statistic}. The final dataset comprises 2,109 videos (0.86M frames total), with lengths ranging from 100 to 3,000 frames. Spanning 419 categories across 4 major underwater scenes, each video contains 16–20 video-language pairs, yielding approximately 40k video-text pairs in total.
\begin{figure*}
    \centering  
    \includegraphics[width=0.95\textwidth]{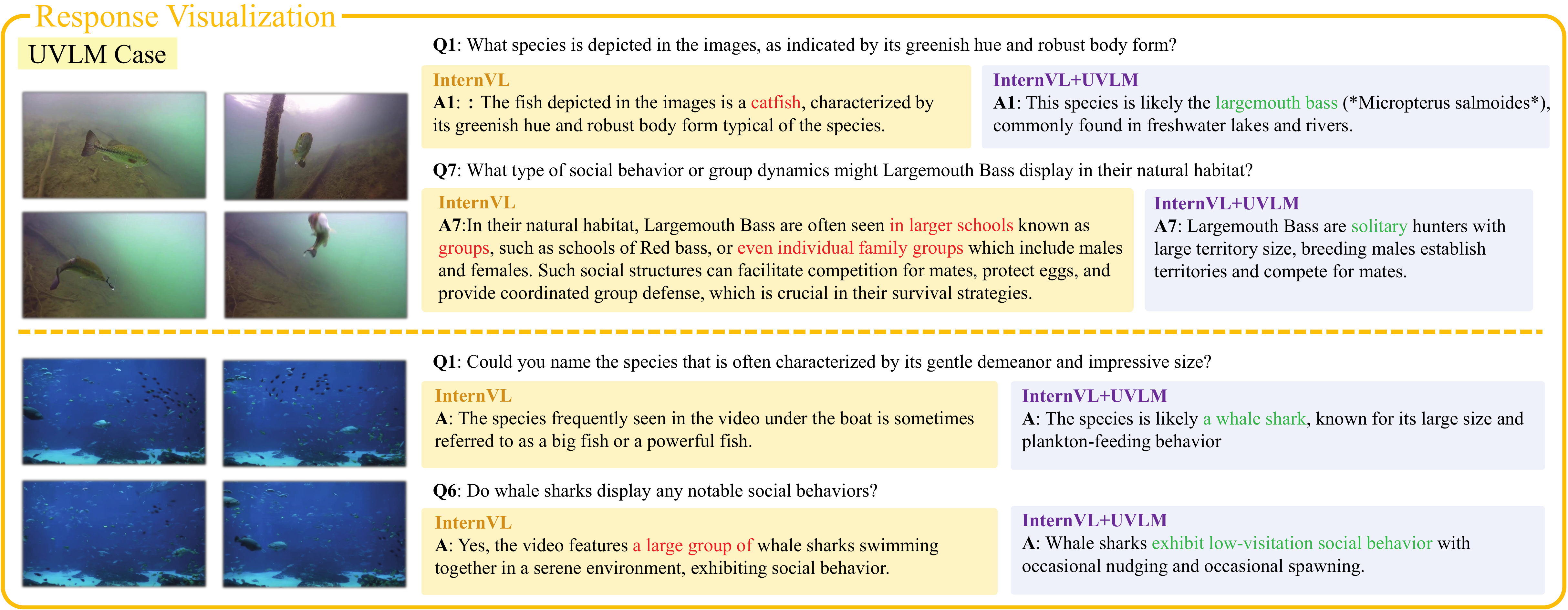} 
    \caption{Visualization examples of fine-tuning with UVLM. Error and correct descriptions are marked in red and green.}
    \label{fig:benchmaik-case}   
\end{figure*}

\subsection{Evaluation Metrics}  


UVLM's comprehensive evaluation employs two Objective Metrics and five LLM-based Judgement Metrics (details in the Appendix.C). Two objective metrics are: 1) Multiple Choice Accuracy, evaluate the model's performance on various common multiple-choice questions; 2) Fine-grained taxonomic classification, evaluate the model's performance on multiple-choice questions related to the professional biological taxonomy~\cite{whittaker1969new}. The LLM-based Judgement Metrics are inspired by LLM judge~\cite{zheng2023judging} and MMDU~\cite{liu2024mmdu}, leveraging GPT-4o as the evaluation backbone. Specifically, they assess model performance from five aspects; 1) Semantic accuracy, assesses how closely the description matches the reference answer; 2) Visual perception accuracy, examines the correctness of the model’s interpretation of image content; 3) Detail completeness, measures the thoroughness of the provided information; 4) Environmental description accuracy, focuses on the correctness of underwater environmental descriptions; 5) Species behavior matching, evaluates the accuracy of prediction. Evaluation prompts are available in Appendix.C.

\section{Experiments}


\subsection{Dataset Partition}
UVLM contains 2109 videos, covering 419 categories. As shown in Figure \ref{fig:statistic} (b), the overall category distribution follows a long-tail distribution and videos of many specific categories have relatively few observed targets. To ensure the training and test sets have consistent category distributions as much as possible, we sampled the test set proportionally. For categories with more than 5 videos, we randomly selected videos for the test set according to a predefined train-test split ratio. Then, we further randomly sampled videos from categories with fewer than 5 videos as test samples. Finally, 208 videos-text pairs were selected as test set.

\subsection{Experimental Results on UVLM}
Table \ref{tab:comparison} and Figure \ref{fig:benchmaik-case} present the results on UVLM. We selected three recently released VidLMs (InternVL~\cite{chen2024internvl}, QwenVL~\cite{bai2025qwen2}, VideoLLaMA~\cite{zhang2025videollama}) and a representative VLM (Qvis~\cite{lu2024ovis}) as baselines.  For InternVL, we evaluated the 1B and 8B variants, while for VideoLLaMA, we tested both the 2B and 7B versions. According to the experimental results, we can draw several observations: 

\begin{itemize}

\item \textbf{Underwater observation poses distinct challenges compared to in-air scenarios}. Even state-of-the-art closed-source models, GPT-4o (77.95) and Gemini (75.00), or open-source models like Qwen2.5VL-72B (75.49), achieve relatively limited performance. The gap highlights unique complexities of underwater scenarios, suggesting significant potential for further exploration


\item \textbf{Fine-tuning with UVLM significantly enhances VidLMs' underwater observation capabilities}. For example, VideoLLaMA3-7B achieves an accuracy gain exceeding 10 points, reaching 73.04, just 2.45 points behind the much larger Qwen2.5VL-72B (75.49). Remarkably, this compact model also closely matches closed-source models like Gemini (75.00), despite their larger scales and proprietary datasets. These results indicate that the proposed human-AI collaborative annotation pipeline effectively distills knowledge from large models like GPT-4o and injects it into smaller models through fine-tuning. While maintaining acceptable performance trade-offs,it significantly reduces model overhead and hardware requirements and expand the model's applicability scope.

\item \textbf{In highly specialized fields, simply increasing the amount of training data is insufficient to bridge the gap between small models and large models}. 
Differing from other six metrics, performance on the fine-grained taxonomic classification task requires complex biological domain knowledge. For this particular task, we observe a persistent performance gap between small and large models even after extensive fine-tuning. We attribute this phenomenon to the fundamental capacity differences between model scales. While simple QA tasks can be effectively addressed through data fine-tuning alone as they impose relatively low demands on model capacity, tasks requiring sophisticated domain expertise present a greater challenge. Small models, constrained by their limited capacity while simultaneously maintaining performance across multiple tasks, demonstrate particular difficulty in closing this performance gap through fine-tuning alone. This finding presents a new challenge for the field of underwater world
understanding.


\end{itemize}

In addition, \textbf{the benefits of UVLM fine-tuning extend beyond underwater observation, improving performance on general benchmarks such as VideoMME and Perception Test}. Additional results are provided in the Appendix.E. 


\section{Conclusion}

This paper introduces the first multimodal benchmark specifically designed to enhance video-language model applications in underwater environments. Combining human expertise with advanced AI annotation methods, UVLM encompasses extensive marine biodiversity (419 animal classes), diverse video resolutions, and realistic underwater challenges, including varying illumination and water turbidity, thus providing authentic conditions for model training and evaluation. Featuring 20 comprehensive subtasks across distinct scenarios with carefully designed evaluation metrics, UVLM enables meaningful quantitative comparisons, fostering the development of accurate and reliable underwater observation systems. We anticipate this resource will promote innovation and contribute to sustainable underwater monitoring and exploration.




\bibliography{reference} 

\appendix
\clearpage
\setcounter{page}{1}
\setcounter{figure}{0}
\setcounter{section}{0}
\setcounter{table}{0}

\section{Appendix Contents \& Benchmark Link }

\textbf{}

\begin{itemize}
\small
    \item A. Sample Generate Prompt Details \& Annotation Error Statistics
    \item B. Task Structure \& Example Visualization of UVLM
    \item C. Objective Metrics  \& LLM-based Evaluation Prompt Details
    \item D. Finetuning Setting Details
    \item E. Evaluation on Generalization Effect of
UVLM Fine-tuning
    \item F. Evaluation with Different Judgment Models
    \item G. Evaluating How Underwater Image Quality Influences Model Accuracy 
\end{itemize}



\begin{figure}
  \centering  
  \includegraphics[width=7cm, height=14cm]{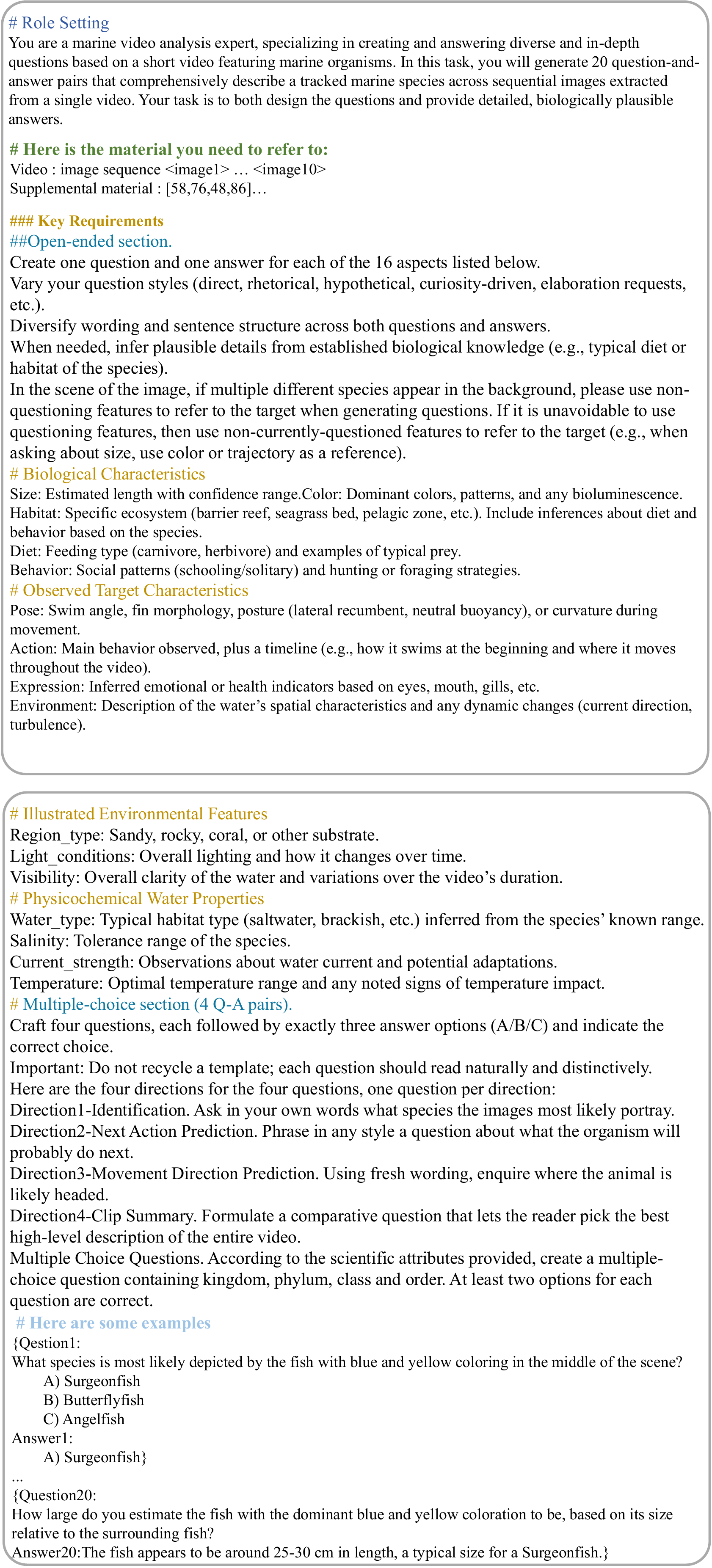}
  \caption{Dialogue generation prompt}
  \label{fig:genenate1}
\end{figure}

\section{A. Sample Generate Prompt Details}

In this section, we provide the prompts we used to generate the samples, as shown in Figure~\ref{fig:genenate1}.

First, our prompt specifies the domain to which the video belongs and the coordinate location of the target object, guiding GPT-4o to pose insightful and relevant questions aligned with the visual theme. We leverage GPT-4o to generate a large number of questions centered on the underwater world, ensuring that these questions comprehensively cover the 20 thematic domains we have proposed. Secondly, to ensure fairness in data generation, we do not provide any textual cues when generating answers. Instead, GPT-4o is required to understand and respond to multiple questions based solely on the video content.

In the initial part of the prompt, as shown in Figure~\ref{fig:genenate1}, we assign GPT-4o the role of a marine video analysis expert, explicitly framing the task as designing and answering diverse multi-turn questions for a video featuring marine life. This role assignment encourages the model to draw from disciplinary frameworks such as marine biology, ecology, and behavioral science, ensuring that the output remains coherent and scientifically grounded. Immediately following this, the prompt provides an overview of the reference materials, including annotated video frames and supplemental cues. These materials serve as core visual and contextual anchors, implicitly guiding the reasoning process behind the generated questions and answers.

The main body of the prompt outlines the key requirements that the model must follow when generating questions and answers, ensuring comprehensive coverage of marine organism research. These requirements are structured into several subfields:
\begin{itemize}
\item
\textbf{Biological Characteristics:}
The model is expected to observe and infer the organism’s body size, coloration, habitat, dietary habits, and social behaviors, as well as possible predation or foraging strategies.
\item
\textbf{Observed Target Characteristics:}
This focuses on the organism’s specific behaviors, swimming posture, action sequences, and any signs of health or emotional state.
\item
\textbf{Illustrated Environmental Features:}
The model should describe the environmental conditions depicted in the video, such as substrate composition, water clarity, and lighting.
\item
\textbf{Physicochemical Water Properties:}
This includes attention to water chemistry and physical parameters (e.g., salinity, temperature, current strength), allowing for deeper inference into species adaptability and ecological preferences.
\end{itemize}
These structured guidelines require the model to integrate information across biological, ecological, and environmental domains, resulting in questions and answers with greater scientific depth and contextual richness.


\begin{figure*}
  \centering
  \includegraphics[width=1\textwidth]{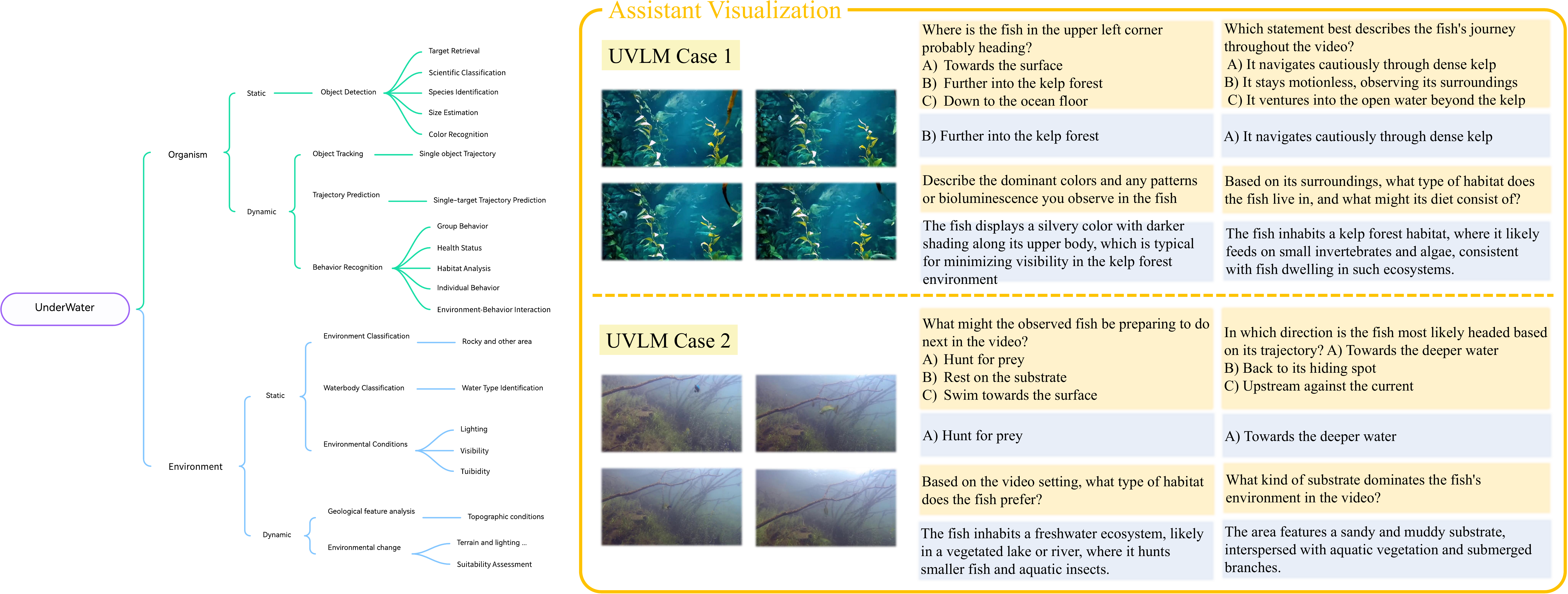}
  \caption{Task structure and the examples of UVLM. Left side: Task structure; Right side: several cases}
  \label{fig:Assistant}
\end{figure*}

\begin{table}
\centering
\small
\begin{tabular}{lcc}
\toprule
Method & Acc\_obj (\%) & Acc\_subj (\%) \\
\midrule
GPT-4o   & 45  & 47  \\
+PK      & 91  & 95  \\
+PK+MC   & 100 & 100 \\
\bottomrule
\end{tabular}
\caption{
Annotation accuracy on objective and subjective labels under different settings.
PK: prior knowledge; MC: manual correction; Acc\_obj/Acc\_subj (\%): annotation accuracy on objective/subjective labels.
}
\label{tab:annotation-accuracy}
\end{table}

\textbf{Annotation statistical reliability:} As summarized in Table~\ref{tab:annotation-accuracy}, incorporating prior knowledge (+PK) boosts GPT-4o from 45\%/47\% to 91\%/95\% accuracy on objective/subjective labels, indicating that most errors stem from missing domain priors rather than random noise. Adding manual correction (+PK+MC) further removes all residual errors on the curated set, yielding 100\% accuracy for both label types. During correction, 9.5\% of videos were discarded due to irreconcilable ambiguity or low visual quality; thus, the true accuracy on the original pool is conservatively bounded between 90.5\% and 100\%. Inter-expert agreement on the remaining items ranges from 91.67\% to 100\% for two-expert groups and from 78\% to 100\% for three-expert groups, demonstrating high annotation consistency.

\section{B. Task Structure \& Examples of UVLM}

The task structure and several cases of UVLM are shown in the left side and right side of Figure~\ref{fig:Assistant} respectively. In UVLM case 1, a fish in the upper-left corner of the frame becomes the focal point of attention. By observing its steady swimming path and the surrounding environment, it can be inferred that the fish is slowly moving deeper into the kelp forest area (Option B), demonstrating a high degree of environmental adaptation and a clear exploratory intent. Throughout its journey, the fish remains within the cover of the kelp, avoiding open waters and showing no significant changes in depth. This cautious movement strongly suggests that its behavioral pattern aligns with "navigating cautiously through dense kelp" (Option A).

Visually, the fish features a silvery body tone with a slightly shadowed dorsal area. This coloration provides natural camouflage in the dim kelp forest, helping reduce predator detection and enhancing foraging efficiency. The surrounding environment appears stable and characteristic of a typical kelp forest habitat, eco-complex and suitable for the survival of various small organisms. Based on this, it is inferred that the fish is omnivorous, primarily feeding on small invertebrates and algae, which is a common dietary structure within kelp forest ecosystems. Taken together, the four interrelated questions not only reveal the fish’s movement direction and behavioral patterns but also connect its physiological traits with its ecological adaptation strategies. This makes the sequence a representative example of fish behavioral ecology within a kelp forest environment.

\begin{figure*}
  \centering
  \includegraphics[width=0.85\textwidth]{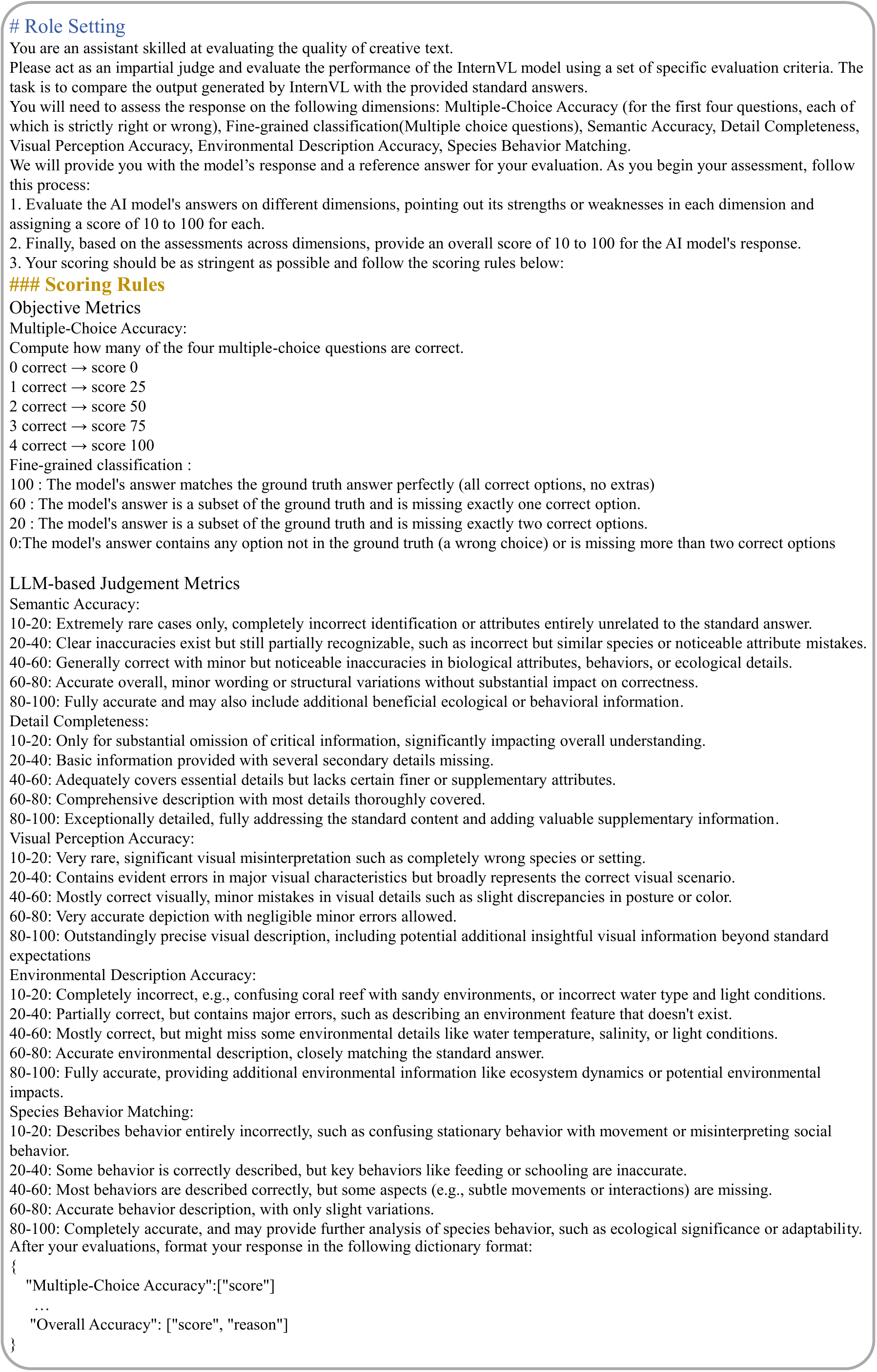}
  \caption{Judgment prompt.}
  \label{fig:eval-prompt}
\end{figure*}

\begin{figure*}
  \centering
  \includegraphics[width=0.9\textwidth]{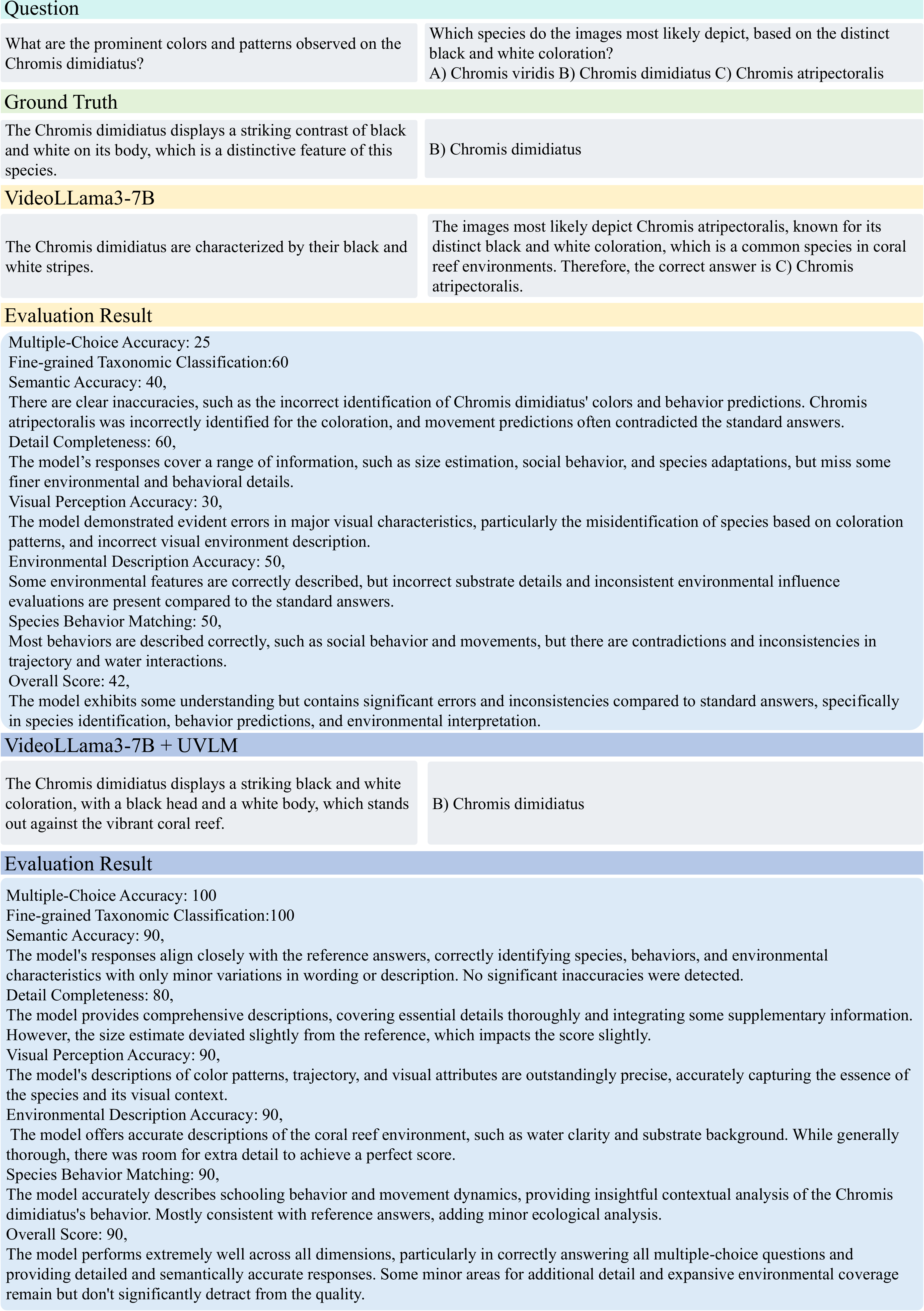}
  \caption{Several evaluative demos of UVLM.}
  \label{fig:UVLM-eval-sample}
\end{figure*}

\begin{table*}[]
  \caption{Representative video language benchmarks}
  \label{tab:VideoBenchmarks}
  \resizebox{1\textwidth}{!}{%
  \centering
  \begin{tabular}{lll}
    \toprule
    \textbf{Name} &  \textbf{Venue}&  \textbf{Themes \& Scenarios} \\
\midrule
MSVD-QA~\cite{xu2017video}, Next-QA~\cite{xiao2021next} & MM$\prime$17, CVPR $\prime$21 & Daily life \\
ActivityNet-QA~\cite{yu2019activitynet} & AAAI$\prime$19 & Human activities in daily living \\
EgoSchema~\cite{mangalam2023egoschema}  & NeurIPS $\prime$23 & Life and human action, egocentric perspective \\
Perception Test~\cite{patraucean2023perception}  & NeurIPS $\prime$23 & Indoor scenes (living room/kitchen), with a small number in bathroom or outdoor settings \\
Video-Bench~\cite{ning2023video} & arXiv $\prime$23 & Television, music videos, NBA highlights \\
InternVid~\cite{wang2023internvid} & ICLR $\prime$23 & People and blogs, education, news and politics, how-to and style, entertainment, gaming, science, sports \\
MovieChat-1K~\cite{song2024moviechat} & CVPR $\prime$24 & Movie clips \\
MVBench~\cite{li2024mvbench} & CVPR $\prime$24 & Life, human action, movies, people, pets, science, instruction, finance, news, sports, humor, etc. \\
LONGVIDEOBENCH~\cite{wu2024longvideobench} & NeurIPS $\prime$24 & Life, movie, knowledge and news \\
E.T. Bench~\cite{liu2024etbenchopenendedeventlevel} & NeurIPS $\prime$24 & Indoor activities, tabletop, sports, egocentric views, cooking, news and vlogs, how-to, open-ended \\
LVBench~\cite{liu2024bench} & arXiv $\prime$24 & Sports documentary, event record, lifestyle, TV show, cartoon \\
EgoThink~\cite{cheng2024egothink} & CVPR $\prime$24 & Daily life understanding from first-person perspective \\
Video-MME ~\cite{fu2024video} & CVPR$\prime$25 & Multilingual, life record, artistic performance, sports competition, film and television, knowledge \\
    \bottomrule
  \end{tabular}
  }
\end{table*}

\section{C. Objective Metrics  \& LLM-based Evaluation Prompt Details}

In Figure~\ref{fig:eval-prompt}, we present the evaluation prompt used to direct GPT-4o in rigorously scoring model’s responses against ground-truth answers. The assessment framework evaluates performance across seven key dimensions, including two objective metrics (multiple-choice accuracy and fine-grained classification) along with five LLM-based judgment metrics (semantic accuracy, detail completeness, visual perception accuracy, environmental description accuracy, and species behavior matching). Multiple-choice accuracy is determined by converting performance on four single-answer questions into a percentage-based score. The fine-grained classification component specifically assesses knowledge of aquatic organism phylogenetics through multiple-choice questions covering four taxonomic levels (kingdom, phylum, class, and order), with strict scoring rules: each omitted correct option deducts 40 points (one omission) or 80 points (two omissions), while any incorrect selection results in an automatic zero for that question. For these two metrics, the LLM serves solely as a statistical tool to measure the discrepancy between model responses and ground truth, ensuring the evaluation is objectively calculated. For the five LLM-based metric, each metric is scored on a 10 to 100 scale using detailed evaluation criteria. For the five LLM-based metrics, performance is measured across five progressive scoring tiers (10-20, 20-40, 40-60, 60-80, and 80-100) with increasingly stringent requirements. This comprehensive scoring system ensures rigorous evaluation across all assessment dimensions. 
The judge must justify every sub-score with concise evidence-based reasoning, then derive an Overall Score that synthesises the dimension-level results.


To clearly illustrate the testing and evaluation of UVLM, we present several Q\&A pairs from UVLM in Figure~\ref{fig:UVLM-eval-sample} Due to space limitations, we are unable to show complete conversations; therefore, each case includes only two representative Q\&A pairs: one multiple-choice question and one open-ended question. Each case is accompanied by the corresponding video and the ground truth answers. In addition, we list the results of both VideoLLaMA3-7B and VideoLLaMA3-7B+UVLM. We also provide evaluation scores generated by GPT-4o, including the reasoning behind the scores and the specific ratings.

\section{D. Finetuning Setting Details}
In the experimental section, we finetuned the InternVL2.5 and VideoLLaMA3 models using our self-constructed UVLM dataset. The detailed evaluation metrics are presented in Table~\ref{tab:Training Set}.
It summarizes the key hyperparameter configurations used during the training of the two multimodal models: VideoLLaMA3 and InternVL2.5. The two models share most of the settings, including a batch size of 128, cosine learning rate scheduling, a warm-up ratio of 0.03, the AdamW optimizer, numerical precision of bfloat16, and the distributed training framework DeepSpeed ZeRO-3.
However, InternVL2.5 differs from VideoLLaMA3 in two notable aspects: it adopts a smaller learning rate and applies a non-zero weight decay.

\begin{table}
\centering
\caption{Training hyper-parameters}
\label{tab:Training Set}
\begin{tabular}{lcc}
\hline
Hyper-parameter        & VideoLLaMA3 & InternVL2.5 \\ \hline
batch size             & 128         & 128         \\
learning rate          & 1e-5        & 2e-6        \\
learning rate schedule & cosine      & cosine      \\
warm-up ratio          & 0.03        & 0.03        \\
weight decay           & 0.0         & 0.01        \\
epoch                  & 1           & 1           \\
optimizer              & AdamW       & AdamW       \\
float precision        & bfloat16    & bfloat16    \\
deepspeed configuration & zero3       & zero3       \\ \hline
\end{tabular}
\end{table}

\section{E. Evaluation on Generalization Effect of UVLM Fine-tuning}
\subsection{Motivation}
Underwater data exhibit a pronounced domain gap relative to conventional terrestrial scenario benchmarks, and training on them directly may introduce capability bias. Fine-tuning on UVLM lets us ask a simple but powerful question: \textbf{Does exposure to a visually and physically distinct domain (underwater) teach a general VLM something that transfers back to everyday terrestrial videos?}

VidLMs are commonly trained on terrestrial footage that implicitly encodes strong priors: gravity points downward, illumination is quasi-static, and hues are only mildly distorted. Underwater scenes violate each of these assumptions. Severe wavelength-dependent absorption, dynamic caustics, six-DoF camera drift, buoyancy-driven motion, and turbidity collectively create an \emph{extreme domain shift}.  
We hypothesize that forcing a VidLM to master these conditions \emph{regularizes} its spatio-temporal representations, yielding better out-of-domain performance. To test this, we fine-tune a general VLM on our underwater dataset, \textbf{UVLM}, and evaluate on three terrestrial benchmarks.



\subsection{Experimental Terrestrial Benchmarks}
Table~\ref{tab:VideoBenchmarks} provides a concise summary of representative benchmarks, highlighting their respective content coverage. Although these efforts have significantly advanced video-language research in terrestrial scenario contexts, the underwater environment remains largely uncharted, motivating our investigation of underwater video-language benchmark. In the following, we provide brief introductions to several well-established video-language benchmarks.

\textbf{MVBench} is a comprehensive benchmark designed to evaluate the performance of large multimodal language models on complex video understanding tasks. It consists of 20 diverse tasks that cover various aspects of video comprehension, including action recognition, object interaction detection, scene change detection, and fine-grained action classification. The benchmark challenges models to process both spatial and temporal information, testing their ability to reason about dynamic content in videos.

\begin{figure*}
  \centering
  \includegraphics[width=0.75\textwidth]{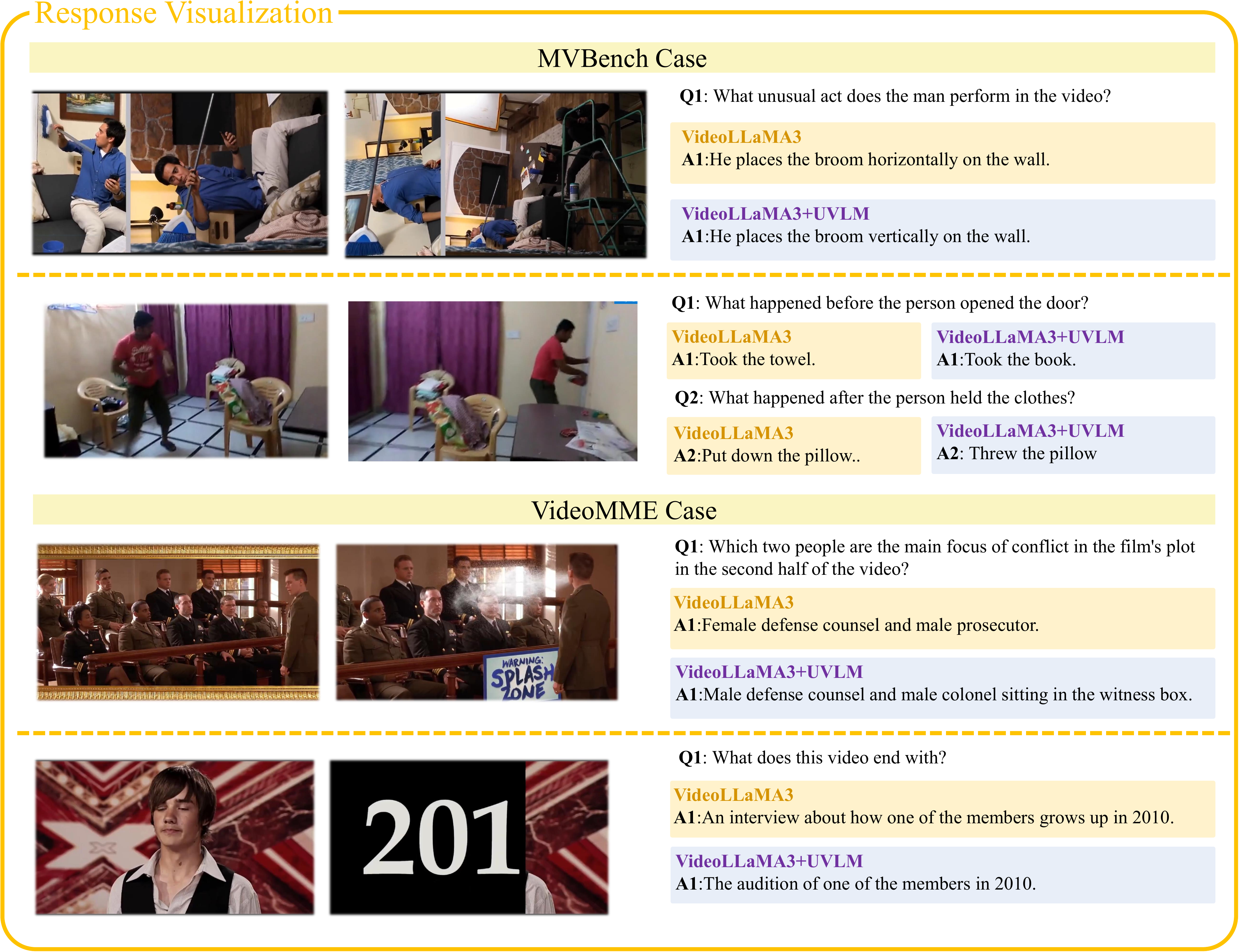}
  \caption{Several demos of comparison on related benchmarks.}
  \label{fig:other-benchmark}
\end{figure*}

The MVBench dataset includes 500 videos from various domains, each accompanied by over 10,000 multiple-choice questions and answers. These questions are designed to assess models' understanding of key elements in video content, such as object presence, action sequences, and contextual reasoning over time. The dataset also includes questions that require models to perform more complex reasoning, such as predicting future actions or understanding interactions between objects within a scene.

\textbf{Video-MME} is a comprehensive multimodal benchmark dataset designed to evaluate model performance on complex video understanding tasks. It has over 900 video clips and 2,700 question-answer pairs. Each video is accompanied by audio and textual annotations, covering a wide range of real-world scenarios. The dataset spans six diverse visual domains: knowledge, film and television, sports competitions, artistic performances, daily life recordings, and multilingual content. These videos also vary significantly in duration, posing challenges for models to process and integrate visual, auditory, and textual information. The benchmark is intended to assess models' ability to recognize fine-grained details, actions, and to understand contextual nuances across modalities.

\textbf{Perception Test} is a multimodal video benchmark designed to evaluate the perceptual and reasoning capabilities of pretrained multimodal models. It focuses on a range of perception skills, including memory, abstraction, physical understanding, and semantics, across video, audio, and text modalities. In addition, it evaluates various types of reasoning, such as descriptive, explanatory, predictive, and counterfactual reasoning. The dataset comprises 11,600 real-world videos and includes six types of annotations, such as multiple-choice questions, video-based Q\&A, object and point tracking, temporal actions, and audio clips.

\subsection{Results and Discussion}

\begin{table}
\caption{Result on other benchmark}
\label{tab:other-benchmarh}
\resizebox{0.5\textwidth}{!}{%
  \centering
\begin{tabular}{lccc}
\hline
Method                & MvBench & Perception Test & VideoMME \\ \hline
InternVL2.5-1B        & 63.2    & 59.8            & 47.7     \\
InternVL2.5-1B + UVLM & 63.3    & 59.6            & 47.9     \\
                      & \textbf{+0.1}    & \textbf{-0.2}            & \textbf{+0.2}     \\ \hline
VideoLLaMA3-2B        & 65.7    & 68.0            & 59.4     \\
VideoLLaMA3-2B + UVLM & 65.9    & 68.2            & 59.2     \\
                      & \textbf{+0.2}    & \textbf{+0.2}            & \textbf{-0.2}     \\ \hline
InternVL2.5-8B        & 72.1    & 68.4            & 59.7     \\
InternVL2.5-8B + UVLM & 72.3    & 68.3            & 60.3     \\
                      & \textbf{+0.2}    & \textbf{-0.1}            & \textbf{+0.5}     \\ \hline
VideoLLaMA3-7B        & 70.0    & 73.0            & 65.6     \\
VideoLLaMA3-7B + UVLM & 70.1    & 73.1            & 65.8     \\
                      & \textbf{+0.1}    & \textbf{-0.1}            & \textbf{+0.2}     \\ \hline
\end{tabular}
}
\end{table}

\begin{table*}[]
\caption{Evaluate the results of the four models using different judgment models.}
\label{tab:Different judement}
\resizebox{1\textwidth}{!}{%
\centering
\begin{tabular}{clcccccc}
\hline
\begin{tabular}[c]{@{}c@{}}Judgement\\ Model\end{tabular}&\multicolumn{1}{c}{Model}   & \begin{tabular}[c]{@{}c@{}}Semantic \\ Accuracy\end{tabular} & \begin{tabular}[c]{@{}c@{}}Visual \\ Perception \\ Accuracy\end{tabular} & \begin{tabular}[c]{@{}c@{}}Detail \\ Completeness\end{tabular} & \begin{tabular}[c]{@{}c@{}}Environmental \\ Description \\ Accuracy\end{tabular} & \begin{tabular}[c]{@{}c@{}}Species \\ Behavior\\ Matching\end{tabular} & Overall \\ \hline

\multirow{5}{*}{GPT4o}                                                           & Claude-3.7-Sonnet                                                 & 73.35  & 73.58      & 74.10  & 79.71 & 76.35 & 76.09       \\
& Gemini-2.5-Flash & 72.43 & 73.34& 70.53  & 78.32  & 74.92 & 75.00 \\
 & GPT4o  & 77.67 & 78.23 & 73.40  & 80.07  & 79.73 & 77.95 \\                 \cline{2-8}
& VideoLLaMA3-7B  & 60.43& 62.05   & 62.07  & 66.97 & 61.41  & 62.70                          \\
& VideoLLaMA3-7B + UVLM & 70.52  & 71.39   & 70.17  & 76.41  & 73.39 & 73.03         \\
  \hline
 
 \multirow{5}{*}{Claude-3.7-Sonnet}
 & Claude-3.7-Sonnet  & 76.82  & 73.90  & 83.70 & 82.91 & 75.55 & 78.99  \\
& Gemini-2.5-Flash                                                  & 79.81& 76.44 & 83.39  & 83.89  & 78.15 & 80.74                          \\ & GPT4o  & 71.78  & 68.41 & 65.94  & 73.10  & 70.67    & 71.92                          \\  \cline{2-8}
& VideoLLaMA3-7B   & 58.65  & 52.93 & 60.26 & 63.12     & 54.06 & 59.5                           \\
& VideoLLaMA3-7B + UVLM & 69.23  & 65.31       & 70.79 & 74.57 & 67.93  & 71.51  \\  \hline

\multirow{5}{*}{Gemini-2.5-Flash}   
 & Claude-3.7-Sonnet  & 81.54  & 80.85  & 83.65  & 90.48 & 79.97         & 82.20                          \\
 & Gemini-2.5-Flash & 83.57   & 82.19  & 82.49 & 91.55  & 83.29          & 83.52                          \\
  & GPT4o   & 76.60   & 73.00& 65.10 & 80.69  & 73.47  & 73.85                         \\    \cline{2-8}
& VideoLLaMA3-7B & 46.93  & 44.21  & 50.98 & 51.84 & 46.90    & 49.59  \\
& VideoLLaMA3-7B + UVLM & 60.84   & 56.54 & 60.36  & 71.42   & 60.77  & 65.57  \\       \hline
\end{tabular}
}
\end{table*}

We report overall performance on these benchmark in Table~\ref{tab:other-benchmarh}. Qualitative comparisons are shown in Figure~\ref{fig:other-benchmark}. Specifically, UVLM significantly enhances the model’s capabilities in spatial reasoning and temporal understanding, enabling more precise and fine-grained comprehension across diverse video scenarios. After fine-tuning with our dataset, the model demonstrates stronger generalization in recognizing object orientation, interpreting target actions, and performing inference. For example, in the first case shown in Figure ~\ref{fig:other-benchmark}, the question 'What unusual act does the man perform in the video?' requires accurate judgment of the object’s spatial orientation. The baseline model incorrectly responds with 'He places the broom horizontally on the wall', failing to capture the actual positioning of the object. In contrast, the model fine-tuned on our dataset correctly identifies the broom as being placed vertically and answers 'He places the broom vertically on the wall', demonstrating a more acute perception of spatial structure.

Our findings suggest a simple yet effective strategy for building universally capable VLMs: supplement standard web video with \emph{challenge domains} exhibiting extreme physics and optics.  
Underwater footage acts as an “immunity booster,” teaching the model to cope with severe domain shifts and improving generalization more efficiently than scaling homogeneous data. Future work will explore other out-of-distribution sources such as infrared, aerial drone, and egocentric streams.

                      

\section{F. Evaluation with Different Judgment Models}


We conduct a comparative analysis of evaluation using three foundational judges: GPT-4o, Claude-3.7-Sonnet, and Gemini-2.5-Flash. By comparing the resulting scores, we seek to verify that UVLM-induced improvements are robust to evaluator choice and to identify any rating bias or self-preference each judge may exhibit. These findings will validate the generality of the fine-tuning effect and guide future work on selecting, weighting, or calibrating evaluation protocols, ensuring fairer and more credible conclusions.

As shown in Table~\ref{tab:Different judement}, integrating UVLM produces clear performance gains across all six subjective evaluation dimensions under every judge, with each judge exhibiting its own preference pattern.

The smallest improvement appears under GPT-4o, where VideoLLaMA3-7B’s Overall score rises from 62.70 to 73.03, a gain of +10.33. Claude delivers the most balanced assessments, assigning Overall scores that lie between those of the other two judges. Gemini reports the largest boost: the Overall score climbs from 49.59 to 65.57, an increase of +15.98. In every case, the introduction of UVLM raises the Overall score by at least ten points, demonstrating robust and consistent benefits across diverse judges and model settings.

Notably, Gemini tends to award higher ratings when it evaluates its own generations: most strikingly, 82.19 for Visual Perception Accuracy, 91.55 for Environmental Description Accuracy, 83.29 for Species Behavior 
Matching, and a lofty Overall score of 83.52. This phenomenon suggests a degree of self-preference bias in Gemini, particularly regarding recognition of visual detail and ecological perception ability.

\subsection{}

\section{G. Evaluating How Underwater Image Quality Influences Model Accuracy}

\begin{figure}
  \centering
  \includegraphics[width=0.48\textwidth]{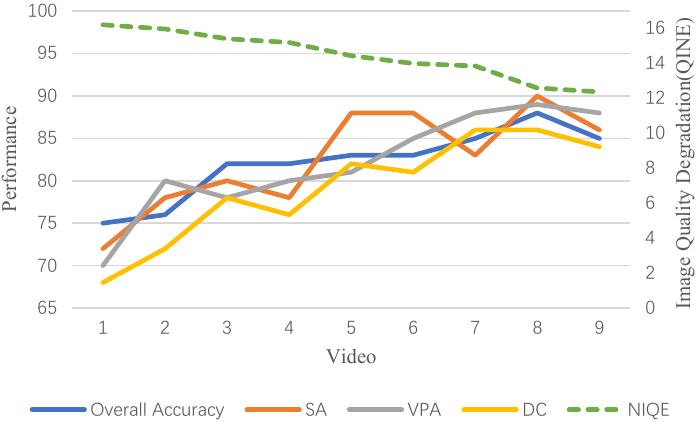}
  \caption{Image quality influence. }
  \label{fig:NIQE}
\end{figure}

To explore the effects of underwater image quality degradation on video perception, Figure \ref{fig:NIQE} presents the image quality degradation curve along with several underwater video perception metrics. Here, we employ NIQE to assess image quality degradation, where a higher NIQE value indicates more severe degradation. Specifically, we randomly selected nine samples. The NIQE score was computed on ten randomly selected frames, and the average of the ten frames was taken as the final NIQE value. For easy comparison, we ranked the videos in descending order of NIQE.

Experimental results demonstrate a clear correlation between improved image quality and enhanced perception performance. As the NIQE scores decreased from approximately 16 (indicating poorer quality) to 12 (better quality), we observed progressive improvements across all evaluation metrics. The overall accuracy increased by approximately 20\%, while semantic accuracy showed particularly significant gains, rising from 70 to 90. Visual perception accuracy similarly exhibited an upward trend with quality improvement.

These findings substantiate the necessity of considering underwater-specific conditions, highlighting the importance of establishing a specialized benchmark for underwater video perception tasks


\end{document}